\documentclass[dvipdfm,a4paper,12pt]{article}
\input{epsf.sty}
\begin{document}
\title{Giving the AI definition a form suitable for the engineer}
\author{Dimiter Dobrev\\
Institute of Mathematics and Informatics\\
Bulgarian Academy of Sciences\\
1113 Sofia, BULGARIA\\
e-mail: d@dobrev.com}
\renewcommand{\today}{April 3, 2013}
\maketitle

\begin{abstract}
Artificial Intelligence -- what is this? That is the question! In earlier papers we already gave a formal definition for AI, but if one desires to build an actual AI implementation, the following issues require attention and are treated here: the data format to be used, the idea of Undef and Nothing symbols, various ways for defining the `meaning of life', and finally, a new notion of `incorrect move'. These questions are of minor importance in the theoretical discussion, but we already know the answer of the question `Does AI exist?' Now we want to make the next step and to create this program.
\end{abstract}

\section*{Introduction}

If you want to answer the question: `What is AI?', the first thing to do would be to consider whom you are giving your answer to, and what is the purpose of your answer -- whether your aim is to convince the reader that AI exists and it possesses certain features that are interesting from a theoretical point of view, or you want to write a guide for designing an actual AI.   

Let's see what happens with a different device, such as the computer. When we look for the definition of `computer', we have to consider whom we are giving it to. Perhaps our readers are mathematicians and are interested in the questions:  `Is the device `computer' actually existing?', `What are its features?', `Could we reduce it to another device that we know? (i.e. reducing the matter to the previous case)'. Mathematicians would be happy to have a definition of computer such as the `Turing machine' or the `unlimited register machine' since what they need is an easy to handle, simple description. Then if you want to undertake a proof by induction, you will check for each and every command in the Turing machine whether it keeps the inductive assumption.  The good thing about the Turing machine is that all commands are of one and the same type. If you try a proof by induction for the model of a real computer, you will face the problem that there the possible types of instructions are in the hundreds and the task becomes next to impossible.

If you want to answer the question: `What is a computer?, and if your answer is intended for engineers, then you will have to say something about CPU and memory, buses for data transmission, the numeral system utilized to encode the data, etc.  

The difference between mathematicians and engineers is first the way they think, and second -- the aim they pursue. For the mathematicians the concept is interesting from a theoretical point of view, while the engineer wants to create a real product and that's why he is interested in a series of technical details.

Here is a typical argument of a mathematician: `I would like to call my friend Peter but I have forgotten his telephone number. No problem, the set of telephone numbers is finite -- I will call all of them and one of them will be the number of Peter.' Such reasoning is very common in mathematical proofs. For the mathematician the question is: `Can I call Peter?' and the answer is `Yes, I can'. Of course, engineers do not use such proofs since they are not interested in the question `Is this theoretically possible?'. Engineers are looking for a real, working solution, even if they might not know why exactly it works.

In articles [1, 2, 4] we tried to catch the fancy of mathematicians, and proposed an AI definition that is useful from a theoretical point of view; however, for the needs of practice, we need to go into details that will be needed for the development of a real program satisfying the definition. The purpose of this article is to catch the fancy of engineers and tell them what Artificial Intelligence is in a way that would be useful to them for the development of a real working program.

In [4] we described one concrete program which satisfies the definition of AI but this program was so inefficient that in order to work it requires infinitely fast computer. In [1] we gave an algorithm which will find an efficient program that satisfies the definition of AI. This is done for finite number of steps but the number is so large that it can be considered infinite. So this algorithm is as useless as the one which answers the question `Can I call Peter?' In the current paper we will not deal with the question about the existence of AI. Here we will turn our attention on practical questions connected with the design of this program.

\section*{Format of the data}

As we've already mentioned in articles [1 - 4], the Artificial Intelligence is a step device that inputs and outputs certain information on each step. Of course, a question arises about the format of this information.

In [1], both the input and the output data were Boolean vectors. Important thing about the AI is the score. This is part of the input which gives us the `meaning of life'.  In [1], the score was expressed by two special bits from the input, which we called `victory' and `loss'. That is to say the data in [1] has some structure, while in [2] the input data are just letters of a finite alphabet (just like the output data). The victory or loss in that case are simply subsets of the set of the input symbols, i.e. the data in [2] has no structure whatsoever.

Article [1] was published in a popular science magazine, which is why it was written for the general public; at the same time, article [2] was designated for a mathematical journal and the technical details were cleared up in order to made the text more precise, while it was assumed that for mathematicians the format of the data is of no importance.

Article [3] reviews a particular world. This is the Tic-tac-toe game. The format of the data is the same as in [1], with the exception that we've added one bit, which we've called `incorrect move'. Article [3] shows that the Boolean vector is not the most suitable format possible, since the input consists of the symbol which is in the current square of the game board. There are three options: empty, circle and cross. Those three options are coded in two bits, where one of the four combinations of the two bits is simply not used, i.e. this input never comes. The situation with the output is similar. The possible actions there are six, which are coded in three bits, where two of the eight combinations are simply not used, i.e. when the device tries to play such combination, it always receives a reply `incorrect move'.

The fact that through the use of coding we can convert data from one format into another means that, from theoretical point of view, the format of the data is irrelevant, but from a practical point of view it is important for us to choose the correct format to avoid the need of coding. The aim of AI is to understand the world; therefore, to make this aim easily achievable, it is good for this world to be simple and easy to understand. If we put a coding at the input and output, this could make the world so complex that our device would not be able to understand it.

Let's take for example the digits from 0 to 9. What will happen if we shuffle them? Well, nothing special.  If we've managed to learn them in this sequence, we will be able to learn the new sequence as well. Let's now take the numbers from 0 to 99 and shuffle them. Let's start with a small shuffle, by exchanging the the first and second digit of each number. This is not problematic since once we've learned to write the numbers from left to right, we will be able to learn to write them from right to left. However, let's apply to the numbers from 0 to 99 a completely arbitrary permutation. Now, that would be a serious problem because we will ruin the natural logic these numbers follow. Let the new sequence be the following:  38, 12, 76, etc. There is no logic at all in this new sequence, and if we want to learn to count from 0 to 99 this would be a serious challenge.

The situation with the Boolean vectors is similar. We are making the presumption that the world is presented in a natural way by them and an eventual coding would complicate the world and will make the task of understanding it more difficult. If we shuffle the naught and the one, this will not be a problem provided that these are two symbols coded arbitrarily by 0 and 1. However, if we assume that these symbols follow a certain sequence (i.e. 0 is smaller than 1), then the shuffling could complicate the world. If we exchange the places of the coordinates in the vector, that would also not be a problem provided that there is no logic in the arrangement of the coordinates, but if the neighboring coordinates are more related than the further ones, then the shuffling could also complicate the world. If we apply an arbitrary permutation to the vectors, this would most surely be a problem because it will hide the logic, according to which the format has been constructed (if there is such logic).

As we've already mentioned in [5], we think for the world as a sum of different factors. Those factors could be independent or could affect one another. For such a world, it is particularly suitable to use vectors to represent data. Vectors will be used to store the internal state of the world as well.  (The world is external to our device. We, the developers of the device, are not concerned with a specific format for representing of the internal states of the world. Anyway, the device will search for a model of the world and in this world model the internal states of the world need to be described. For this purpose the vector format is very suitable since if we have two independent models of the world we can easily unite them into one where the new states will be a concatenation of the vectors of the states of the two models.)

\section*{Signals}

Definition: A function of a single argument (the time) that returns a scalar is called a signal.

This means that the coordinates of the vectors are signals. Let's take four types of signals: input, output, score and internal. The output signals are the coordinates of the output vector. The input and score signals are the coordinates of the input vector.

Note: We've explicitly separated the input in two: score part, which gives us the meaning, and a purely informational part. Same thing was done in [1] where the score was expressed by two signals, which we called `victory' and `loss'. The same thing was done in [3], where we introduced a third score signal called `incorrect move'. The situation in [2] is quite different -- there we have only one input signal that encodes both the information and the score part of the input.

Apart from the signals taking part in the input and output vectors, we also have a lot of internal signals, which the device needs in order for constructing a model of the world. We will assume that the device is looking for a representation of the world by a vector, whose coordinates are some of the internal signals. The set of all internal signals possible is infinite but at any specific moment the device has concentrated on a finite number of internal signals that it finds interesting and adequate to the world it has fallen into.

\section*{Non-Boolean vectors}

Well then, once we've decided that the data and the internal states of the world will be stored in a vector format, let's see if we could limit that format to the Boolean vectors only. The answer is `No'. In [3] we saw that if we use only Boolean vectors, this acquires additional coding.

If we limit the set to the Boolean vectors only, the input and output signals will be Boolean functions. Let's expand the definition with some more complicated signals. We allow for the signal to return $k$ possible values from the $\{0, 1, ... , k-1\}$ set instead of having only two possible values from the $\{0, 1\}$ set. We assume that the sequence of these values is not an arbitrary one (i.e. that the `bigger' order in the $\{0, 1, ... , k-1\}$ sequence corresponds to some natural order typical of the world, if, of course, such an order exists). We assume that because we take the world to be represented in the most natural way possible without requiring additional re-coding. 

Thus, we've expanded the set of signals to the set of finite functions. We will expand it a bit more by allowing the signal to return infinite scalars as well, in the form of an integer, a natural or a real number. We will again assume that the presentation of the data as an integer or a real number is not arbitrary but connected to the internal structure of the world. Therefore, we will expect for the real numbers to obey the property of continuity (i.e. that small changes will not have a material significance).  We will expect that the `bigger' relation is not an arbitrary one as well but related to the natural structure of the world.

\section*{Representing infinite objects as finite} 

We will consider a device with vector input and output, with scalar vectors, where some of the scalars are finite but there could also be infinite scalars. But we need a practical solution where we have a program that inputs and outputs a finite quantity of information at each step. The assumption that we can have countable scalars leads to the result that the input and output will not be finite. If we also assume uncountable scalars (such as the real numbers) this leads to the result that the input and output vectors cannot be coded at all as a finite sequence of bits.

This means that the allowance of infinite objects as a part of the input and output leads to a theoretical model that does not completely correspond to the practice. However, we can assume that the natural and real numbers taking part in the input and output vectors are not real but computer-represented (for example, they are stored in 64 bits, coded with the use of the standard computer coding). Thus we have two models. The first one is theoretical; in it the device operates with an input and output that contain real and natural numbers. With the second model, those numbers are not real and natural but rather the pseudo-real and pseudo-natural ones the computer operates with.

Which of those two models will we work with? We will work with the theoretical one and will apply the results in practice using the practical model. That's the same thing people dealing with Turing machines do. They work with the theoretical model of a computer with an infinite memory (since the Turing machine uses an endless tape). Then they apply the results to real computers that have finite memory.

The only case we will use the practical instead of the theoretical model would be when we want to use the fact that the sets of the inputs and outputs are finite. We will need this to produce some otherwise pointless evidence of existence of the type: `Telephone numbers are finite in number, therefore I can call Peter.'

\section*{The Undef and Nothing symbols}

Many programming languages introduce a special symbol for the case when we do not have a value or we don't know the value.  This symbol will be useful in our case too. We will call it the Undef symbol.

Another special symbol will be Nothing. We will need it for the case when nothing happens. We will not give a specific definition of a situation when nothing happens in the world, but will consider that one of the input symbols has an additional meaning and that this is very informal and for reference only, without having any concrete meaning.  For example in the real world, one could posit silence to be noted with symbol Nothing. It is interesting that when someone sleeps in a noisy room, he wakes up if the noise suddenly stops. That means that the sudden silence is perceived as an interesting event. This means that the symbol Nothing is a symbol as all other symbols and its meaning as `nothing' is for reference only. 

Is it necessary for a specific symbol to have a special meaning? From theoretical point of view it is not, but in this article we deal with the question of the practical solution of the problem, and we want to help the device to understand the world to a maximum extent. That's why we presume that we've provided it with some preliminary information. This information is the special meaning of the symbol Nothing.

When will we use the symbol Nothing? At the input stage for example, we will use Nothing when there is no input. Of course, the unavailability of an input is also an input, but this is a more special input.

When will we use the symbol Undef? This will be when the input is unknown (for example, when a sensor failed or delayed the transmission of the information). We can even assume that there is a chance for the missing information to appear with a delay (i.e. that after one more step certain information will come that the value of Undef at the previous step should have been this or that). Of course, our device should be able to receive and process such information because otherwise that information would be useless.

From any signal we could derive a new one by saying `that signal before $k$ steps', i.e. this is a memory of the signal. There is, of course, the question how to define this memory for the moments when $t-k$ is a negative value, i.e. what to remember for a moment before the birth. The natural value suitable for this case is the symbol Undef.

Let's look at another internal signal. Let's have a model of the world consisting of a finite-state non-deterministic automaton. Let's have a signal bringing back the state of this automaton in the t moment. For a deterministic automaton we would know its respective signal, but for a non-deterministic one there might be moments when we don't know the current state it is in. Then it would be natural for the signal to return the Undef symbol. In a few more steps we may find out what state it was in. Then, certain additional information might come in, telling us that the signal from that step has been this or that rather than Undef.

As you can see, the introduction of symbols Undef and Nothing is very suitable for the input signals and even more suitable for the operation of the internal signals.

With output signals we will also use the symbol Nothing. Imagine that at a certain moment you don't know what to do. In this case, it would be most suitable to do nothing but you have to do something because your device must come out with an output because otherwise it would stay stuck in silence, which, we assume, is unacceptable.

Which should be the output corresponding to `I'm doing nothing'? Let this be the symbol Nothing. Any other symbol could play this role but we assume that we've simplified and standardized the world to a degree that makes it easier to be understood.

It is natural for the symbol Nothing to be the anchor for the device to hang on to when it comes to (gets born in) a completely unknown world. Its first step would be to see what will happen when it plays Nothing along all coordinates of the output vector. Then it will try to give value to one of the coordinates and leave the others be Nothing.

We could assume that the symbol Undef can also be used in output signals, proposing that after a delay of a few steps the device has the right to specify the value of the output and say what should have been in the place of the symbol Undef. This assumption would make things very complicated and that's why we are not making it. The symbol Undef would not participate in output signals.

We do need the symbol Nothing in score signals. In articles [1, 3, 4] we assumed that the score is made by two Boolean signals called `loss' and `victory'. When those two signals simultaneously have the value of 1, we take this for a draw, and when the two signals simultaneously have the value of 0, we assume that we have no score. As you can see, there is one redundant coding added. In [4] we calculated the success of the device by calculating the arithmetic mean of victories, losses and draws, but this arithmetic mean does not include the cases where we obtained no score at all.

It is not logical to assume that our device will get a score at each and every step. It is better to assume that most steps do not get a score, and therefore we will use the symbol Nothing in those cases.

To make things simple let's assume that the score is given by one signal which has a value of a 1, 0 or 1/2 respectively in the cases of victory, loss or draw, and in cases when there is no score, the signal will have a Nothing value. Then the success of the device will be the arithmetic mean of the values of this signal which are different from Nothing.

To make things simpler, we will assume that for the cases of input, output and internal signals the symbol Nothing coincides with the symbol zero. We decided that one of the symbols will have a more special meaning and that the most suitable symbol of the purpose is zero. The other advantage is that it is present in all formats. It is present in the Boolean format, in the finite format, in natural numbers, as well as in real numbers, etc. 

The only place where we assume that zero and Nothing are different symbols is in the score signal. There we make the arithmetic mean of all values different from Nothing, and the zero is a normal number, which can easily participate in the arithmetic mean. To underline that Nothing does not participate in the calculation of the arithmetic mean we will assume that it is not a zero but something different.

\section*{Possible scores}

In [1] we said that a program recognised as Artificial Intelligence is any program which copes better than a human being in an arbitrary world. To be able to compare and say who did better and who did worse, we must have an order concerning lives telling us which life is better and which life is worse.

We will call this order `the meaning of life'.We will first define it for the case of finite life, and then we will expand it for the case of infinite life.

Let's take an arbitrary linear order of finite lives (Let us remind you that we call life the sequence of input and output vectors from the moment of birth until a given moment in time or to infinity).

Each linear order between finite lives has a corresponding evaluation function, which we will call Success function. If one life is better than another, the Success function for the better life returns a bigger number.

Now, lets expand the definition of meaning of life so that it includes infinite lives in a natural way. Let's take a sequence of the beginnings of an infinite life. Let's calculate the value of the Success function for each beginning. We get a sequence of real numbers and if this sequence is convergent, it is natural to define the Success function value for this infinite life to be equal to its limit (plus and minus infinity are also possible values). If the sequence is divergent, then we will consider that the Success function for this infinite life does not have an exact value but that its value is somewhere between the limit inferior and limit superior of the sequence. Thus, we get a new Success function, which returns an exact value for some lives, and an interval of the lowest possible to the highest possible value for other lives.

The new Success function defines a partial order in the set of all lives (this order will not be linear). One life is better than another if its Success function value is bigger, or if an interval is returned -- if that interval is to the right from that of the other life.  

Artificial Intelligence is a device left to understand the world on its own; however, the meaning of life should be given a priori. We cannot expect the device to cope well, if it does not know which result is good and which is bad.  The device must be able to calculate the Success function on its own at any given moment, and the function must depend on the score signals only. The input and output signals supply information about the world situation but do not point towards the meaning of life.

The simplest solution would be to assume that there is just one score signal returning the value of the Success function. This solution is not very good however because in this case the world has to store information about the history of the device up to the present moment so that it can give as score an overall evaluation of its entire life. It is better to have one score signal and the Success function to be the arithmetic mean of all values of this signal. Thus the world will evaluate the device for its last step and not for its entire life up to the present moment. When nothing interesting has happened on the current step and there is no need for the Success function to change, we could assume that the signal returns the previous value and thus the arithmetic mean will be preserved. But in this case the world has to keep track of what was the value at the step before. That's why for this case we will return the symbol Nothing. This is a simpler way to preserve the arithmetic mean.

Whatever the Success function is, we can find a score signal whose arithmetic mean is exactly this function. This score signal is not fully specified because returning the symbol Nothing cannot be distinguished from returning the current arithmetic mean. We should also point out that we assume that for the empty life (life with a length of zero) the Success function returns zero. This is not an issue because if we add a constant to the Success function or multiply it by a positive constant, we will not change the order it defines.

Therefore, an arbitrary meaning of life could be presented as the arithmetic mean of an score signal which returns a real number. Nonetheless, we do not like this solution because we would like for the world to be stable and the evaluation (the Success function) not to be able to jump uncontrollably. That's why we will assume that the score signal is a finite function and returns a value from the set $\{Nothing, 0, 1, ... , k\}$, i.e. we will assume that it could take $k+2$ possible values. The Success function will be in the $[0, k]$ interval. 

This solution is also not a perfect one because we might want to have different levels of priority. For example, it is important not to be late for school but it is much more important not to die in a car accident. Here, `much more important' means `infinitely more important'. We want our definition to enable the world to have N levels of priority. For the purpose thereof, we will assume that we have N score signals and that the Success function returns not a number but a vector. This vector is derived by calculating the arithmetic mean by coordinates, where for each coordinate the sum is divided by the number of times this coordinate has been different from the symbol Nothing. The comparison between two such vectors will be made coordinate by coordinate. We take the coordinate of the highest level of priority, and if for this coordinate the values of the two vectors are equal, we take the next coordinate and so on and so forth.  

Could we emulate a world with two levels of priority, when the score signal is not limited? Yes, let the small priority return 0 or 1, and the big priority return 0 or 2 multiplied by the number of times up to the current moment when the value of the signal has been different from the symbol Nothing. In addition, the world must remember the moment when it came out with 2 multiplied by something, and from that moment on add 2 to each score. Thus, if a score is obtained only from the small priority, the Success function will be in the interval [0, 1], but if we have a score from the high priority, the Success function will be bigger than or equal to 2. Here, with this emulation, we eliminated the limitation of the score signal and burdened the world to memorise what has happened in the past.

Is there a meaning of a life that cannot be presented by $N$ levels of priority? The answer is `Yes'. Let's take a meaning of life that has a countable number of levels of priority. This could be emulated by an unlimited score signal but cannot be presented by $N$ limited score signals for any $N$. Which means that by choosing this definition of the evaluation not every meaning of life would be possible. However, we believe that the worlds with $N$ levels of priority are sufficient in practice, and that it is not necessary to work with worlds with countable levels of priority. What is more, for practical work, in most cases, one level of priority is sufficient. 

Apart from the N score signals from which we calculate the Success function, we will also have another Boolean score signal, which we will call `an incorrect move'. This signal is discussed in the following section.

\section*{Incorrect move}

What is an incorrect move? For example, in chess, if a player tries to move knight as a queen, this would be an incorrect move. Also in chess, a player may not make any move which places his king in check, i.e. any move after which a player is in check is an incorrect move. There are many games in which capture is compulsory. In games like these, an incorrect move is when a player can capture but does not do so.

It is clear that in most worlds there are incorrect moves. Provided that we've fixed the set of output vectors, it is natural to assume that not all of their values represent a correct move. It is normal, for a given output vector to represent a correct move at a given moment and at another moment to represent an incorrect move. 

To enable the world to have incorrect moves, we should answer two questions: `What happens with the world when the device undertakes an incorrect move?', and `What feedback does the world give to the device that the last move was an incorrect one?'

We do not discuss incorrect moves in articles [1, 2]. In these articles we assume that the world punishes the device for an incorrect move by slapping it on the wrist (i.e. gives it a bad score). For example, in the world where you play chess what will happen when you try to perform an incorrect move? One possible solution is for us to define the world so that with each incorrect move you lose the game and you automatically start a new game.

Article [3] introduces a separate signal called `an incorrect move'. In this article we assume that the attempts of the device to undertake an incorrect move do not lead to a change in the world. The result is that the world remains in the same internal state but returns the signal `incorrect move' to the device. The mistake in [3] is that the incorrect move is taken as punishment and it is assumed that the device will learn to perform only correct moves and will avoid the incorrect ones to avoid being punished. 

The information about which move is correct and which is not is very important for the understanding of the world. Let's take for example the situation when we are trying to find our way in the dark by touching the walls with our hands. The touching of the walls could be taken as an incorrect move because we are trying to push our hand through a space it cannot cross. Nevertheless, we are consciously making this incorrect move in order to find out where the wall is.

Maybe it's better to change the definition of a world given in [1, 2] and add one more function called Correct, to the set of functions World and View. For each internal state of the world the Correct function returns the set of all possible moves.

We want to make the task of the device the simplest possible, and define the world in the most easy to grasp way. That's why, it is reasonable to assume that at each step the device receives as an input not only the value of the function View but also the value of the Correct function.

Thus, we have two problems emerging:

The first problem is that the information returned by the Correct function could be too much. If the possible outputs are $k$ in number, the possible values of the Correct function are $2^k$. Respectively, if the outputs are countably many in number, the value of the Correct function is a continuum, etc.

The second problem is that in this way we will complicate the world by imposing the requirement for it to calculate the Correct function at each step, to encode the result in a suitable format and transmit it to the device. 

We will get rid of these problems, if we assume that the world does not explicitly tell the device which answers are correct but that at each incorrect move it returns information that the move is incorrect.

We assume that we have one Boolean signal called `an incorrect move'. We will allow the device to make incorrect moves and when this signal returns a one, we will not consider this a punishment for the device but a piece of useful information.

Nevertheless, we will make four assumptions:

1. We will assume that the incorrect move does not change the internal state of the world, i.e. by making an incorrect move the device loses nothing. We may say that neither it gains something. The only thing gained is the information it obtains. By making a move and this move turns out to be an incorrect one, the device obtains the information that this move is incorrect, which may turn out a piece of useful information.

2. We will assume that if we've tried one move and the world has told us that it is incorrect, there is no need for the device to try it one more time while the state of the world is the same. Of course, if we assume that the Correct function is fixed and we know which the correct moves are before we've tried them, the above assumption is correct, but we might want to assume something even weaker. For example, imagine a world with a built-in clock, which marks how much time the device took to think. In this world, the Correct function depends not only on the state of the world, but also on the time we took to make a move. Well, we will assume that even if the Correct function changes according to the delay, the incorrect moves will only increase in number, i.e. we will assume that if a move is incorrect, it will be still incorrect even if we repeat it. 

3. We will assume that the device has no right to make the same incorrect moves infinitely, i.e. that it must remember the incorrect moves it made and not repeat them at least until a correct move is received. Once a correct move is received, the device may clear its memory and may attempt moves that have been incorrect before. The fact that a move has been incorrect the previous step does not mean that it will be incorrect the next step. 

The reason we made the aforementioned assumption is to prevent the device from reaching a deadlock. Of course, the possible incorrect moves could be many or even infinitely many, which could again result in a delay or deadlock, but knowing that the possible outputs are finite or pseudo-infinite (which is also finite), we will come to the conclusion that, at least in theory, such a deadlock cannot occur.

4. We will assume that the Correct function never returns the empty set, i.e. we will assume that there is always at least one correct move. If we assume that dead ends exist (i.e. situations in which the Correct function returns the empty set), we can associate these moments with death. We can think of death as a mistake which we are trying to avoid but this mistake is always fatal, and therefore we cannot learn from it. That's why it is better to think that there are no such moments in our world.

And yet, when we are programming the device called AI, we may program it to look for a state in which there are more possible moves. In chess, we are trying to deploy our figures so that the available moves are as many as possible. The loss of the queen highly reduces the number of possible moves, which makes this loss unwanted. In life, people strive for freedom. This means they want to have as many possible moves as possible. Any person closed in a very little or narrow place feels uncomfortable. Any locked or shackled man feels uncomfortable too. That's how we can explain people's strive for money and power because this gives additional freedom. When you have money you may or may not buy a boat, but if you don't have money, you don't have a choice. Therefore, man instinctively strives for a state giving him more opportunities. It is logical to use this principle with Artificial Intelligence. If our device avoids cases when the possible moves are few in number, it tries to avoid death as well, since it is the case when we have no possible moves left.

\section*{How will we use the incorrect moves?}

OK, our device understands the world and knows which move is correct and which not. How will we expect it to act? When one move is incorrect with certainty, the device will not attempt it, in order to avoid losing processor time (here `with certainty' means with a very high probability because nothing is absolutely certain). When a move is incorrect almost certainly, the device will try it because it loses nothing by trying it but only obtains information. If the move is really an incorrect one, this will be known with a much greater certainty the next time and, if by chance it turns out to be correct, the device might lose but it might also win by finding new unsuspected opportunities. When it doesn't know whether the move is correct or incorrect, the device might try it but may also not try it. On one side, it will want to check whether the move is correct, but on the other, it will fear eventual unpleasant consequences. For example, you are not trying to jump off the window to see whether you will manage to do it, because if you do succeed by chance, the consequences might turn out to be very bad.

Question: Are incorrect moves part of life? When we are trying an incorrect move, does this increase the number of steps (i.e. the parameter of time)? The answer is `No'. If we look at the Boolean signal `incorrect move', we will see that it is zero for each t, i.e. all moves saved in the history are correct ones. We will assume that the incorrect moves are simply not saved in the history (life).

Life is a sequence of input and output vectors. If the signal `incorrect move' is one of the coordinates of the input vector, this coordinate is always a zero. We will assume that the signal `incorrect move' is not a part of the history because it is useless to include a signal which is constantly a zero.

After all, we said that we want the information obtained from the incorrect moves to be available for use. That's why, we will change the definition of life by inserting the set of vectors of the incorrect moves the device already tried between the input vector and the correct output vector. Life will become a sequence of an input vector, a set of incorrect output vectors, a correct output vector, etc.  

The following article will discuss the dependencies without memory. These are dependencies of the type: `If I see this and do that, the result will be this and that.' These dependencies are represented as implications of the type: $a(t-1)=1, b(t)=0, do(t)=1 \Rightarrow bad\_move(t+1)=1$. This implication must be read as follows: If the signal $b$ at this step is a zero and if the signal $a$ at the previous step was a one and if we chose the signal $do$ at this step to be a one, then this is an incorrect move. We chose the signal $do$ because it is an output signal, and the signals $a$ and $b$ are given because these are input signals which we do not chose but are given by the world.

This implication leads to $bad\_move=1$ but this will not be saved in the history because only the correct moves are saved there. 

The aforementioned implication tells us that in specific circumstances a certain move, will be incorrect. Here we see that on the basis of the information collected from the incorrect moves we can learn to predict such moves. A future publication will show how from the fact that a certain move is incorrect we can extract more information about the state of the world and what would happen at the step after.

\section*{Adding the incorrect moves to the definition of AI} 

In article [4] we've defined AI as a device whose IQ is sufficiently high. The calculation of the IQ uses as a base a set of test worlds by taking the average success of the device for the worlds within that set. The worlds used in [4] are the worlds generated by an arbitrary Turing machine (whose complexity does not exceed one value which is a parameter on which the definition of AI depends).

The set of test worlds we've used in [4] has been selected so as the worlds are maximally natural and comprehensible. The objective is not to encumber the device and make it understand incomprehensible worlds; on the contrary -- we are trying to make things easier for it by making the worlds maximally natural and comprehensible.

One of the problems of article [4] is that there all moves are correct, i.e. the device we've defined in [4] has no idea what an incorrect move is and would not be able to cope in a world in which some of the moves are incorrect ones.  It would be nice if we could adjust the definition in [4] so that it allows worlds with incorrect moves. What is more, by adding the incorrect moves, the worlds would we've chosen for test worlds would become more natural and comprehensible.

One problem in [4] are the worlds that may get into a deadlock. If the world is generated by an arbitrary Turing machine, this machine could at some point find itself in deadlock.  The problems here are two: how do we know that the machine is in deadlock and what shall we do once we know it. The first problem in [4] is solved in a simple way:  if the machine does not come up with a result in 800 steps, we shall consider that it is in a deadlock. We will not change the solution of the first problem, but we shall alter the solution of the second one.

In [4], when the Turing machine is in deadlock we stop it and restart it, i.e. we are saving what has been recorded on the tape, but we are changing its internal state so that its next state is the initial one. However this is not a good solution. It is like unplugging your PC out of the socket without taking care of what is to remain saved on the hard disk drive. An attitude like this towards your PC could make its behavior rather complicated and unpredictable. By shutting-off the deadlocking program we are unduly making the world it generates more complicated. It would be better if once we've stopped it we say to the device that this move has been incorrect and reboot the machine by recovering the information recorded on its tape up to the moment the machine started working on that incorrect move. Thus, the fact that we've made an incorrect move would not have an impact on the future in the respective world.  Thus, the world generated by the machine would be much more natural and comprehensible.

There is yet another problem. It is possible for the arbitrary Turing machine to reach a state (of the tape) where each move leads to a deadlock. This means that we will have to give up the requirement of the existence of at least one correct move. The latter means that the world has to be one in which the device can `die'. In [4] we've already given up the requirement for the world not to contain any fatal errors. Following the same logic and taking into account the same reasons, we could give up the requirement that it is not possible the device to `die'. The notions of `death' and `fatal error' may look synonymous, but if you take a look at our definitions of them, you will see that they are two different things.

In [4], there is one more case when the Turing machine is shutting-off. There the life consist of 100 games each one no longer than 1000 steps. That's why if one game continues more than 1000 steps it is shutting-off. Anyway, this is not a real shutting-off because we only add one ``draw'' score without changing the way the Turing machine works. (We don't change its internal state, nor the configuration on the ribbon.)

\section*{Example} 

We will use the example we reviewed in [3]. This is the world of the game Tic-Tac-Toe, where the device does not see the entire board but only a single cell from it (Figure 1).

\begin{figure}

\begin{center}
\epsfxsize=5cm
\epsfbox{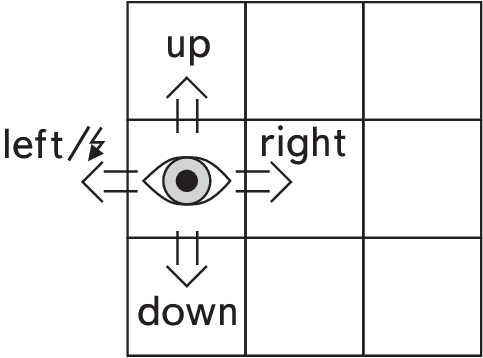}
\end{center}
  \caption{The board}
\end{figure}

The eye of the device is located on top of the cell being viewed. The possible moves are six. The eye may move in four directions, we can put an X in the cell the eye is located on at the moment, and the sixth command is to request a new game (i.e. to clear all cells and start the game anew).  

In article [3] the world used only Boolean vectors; here, we will give up this limitation. This will make the definition of the world simpler and it will become easier to understand.

Instead of two Boolean signals to encode what the eye sees, we will have one input signal with three possible values $\{0, 1, 2\}$, which will correspond to an empty cell, an X and an O. Instead of the two score signals `victory' and `loss', we will have one with four possible values: $\{Nothing, 0, 1, 2\}$, which will correspond to `no score', `loss', `draw' and `victory', instead of the three Boolean output signals, which coded the six possible outputs, now we will have four output signals. The first two will give us the direction of movement of the eye. We will call them $vertical$ and $horizontal$. Their possible values will be in the set $\{0, 1, 2\}$, which will correspond to `does not move', `up' and `down', or respectively `does not move', `left' and `right'. The other two output signals will be Boolean and we will call them $put\_cross$ and $new\_game$. Their functions are clear.

In [3] we had six possible actions and at each move we could perform only one of them. Now we can make four actions at one and the same time. If we want, we can perform no action (by placing a zero on the four coordinates of the output vector). We could assume that we have the right to perform only one action and any other output is taken by the world as an incorrect move, but it is more interesting to assume that we can perform up to four actions in one single move. For example, we can put an X, move up and left and request a new game. Of course, all four actions must be correct because otherwise we will receive `an incorrect move' and nothing will happen.

When we put together several actions in one move, we must specify their sequence. It is all the same whether we will first move up and then left or vice versa.  It is all the same whether we will first move and then request a new game or vice versa. The only action which cannot commute with the rest is placing an X. That's why we will always assume that we have first put the X and then made all other actions.

Thus we present the game Tic-Tac-Toe from [3] as a world of one level of priority, with one input signal, four output and two score signals. (The second score signal is `an incorrect move', which remains as defined in [3].) 

Why is the current representation of the world better than the one made in [3]? Because we have lesser coding, which makes the world simpler and easier to understand. Let's take for example the rule: `If the cell that you see is not empty and if you try to put an X, this is an incorrect move'. Now this rule could be presented as an implication of only three atoms:
$cell(t)\neq0, put\_cross(t)=1 \Rightarrow bad\_move(t+1)=1$

In [3] this implication would comprise six atoms because there the signal cell is encoded with the use of two Boolean signals and the output – with the use of three signals. When we try to find a dependency without a memory, such as the above one, we have to decrease the number of implications by taking only the shortest of them. For this reason, the shorter an implication is, the better the chance for our device to locate it.

\section*{Bibliography}

.

[1] Dobrev D. {\it AI -- What is this}, In: PC Magazine -- Bulgaria, November'2000, pp.12-13 (www.dobrev.com/AI/definition.html).

[2] Dobrev D. {\it A Definition of Artificial Intelligence}, In: Mathematica Balkanica, New Series, Vol. 19, 2005, Fasc. 1-2, pp.67-74.

[3] Dobrev D. {\it Testing AI in one Artificial World}, Proceedings of XI International Conference ``Knowledge-Dialogue-Solution'', June 2005, Varna, Bulgaria, Vol.2, pp.461-464 (www.dobrev.com/AI/).

[4] Dobrev D. {\it Formal Definition of Artificial Intelligence}, In: International Journal ``Information Theories \& Applications'', vol.12, Number 3, 2005, pp.277-285 (www.dobrev.com/AI/).

[5] Dobrev D. {\it Comparison between the two definitions of AI}, In: arXiv:1302.0216, January, 2013.

%%% Author's Address(es); if more than one authors,
%%% addresses are numbered like authors in beginning of paper

\end{document}